\documentclass[10pt,letterpaper]{article}

\usepackage{cogsci}

\cogscifinalcopy %

\usepackage{pslatex}
\usepackage{apacite}
\usepackage{float}
\usepackage{graphicx}
\usepackage{subcaption}
\usepackage{listings} 

\usepackage{xcolor}
\usepackage{amsmath}
\usepackage{amssymb}

\newcommand\E[0]{\mathop{\mathbb{E}}}

\newcommand\codePlaceholder[0]{$\blacksquare$}

\title{Program-Based Strategy Induction for Reinforcement Learning}

\author{{\large \bf Carlos G. Correa$^1$ (cgcorrea@princeton.edu)} \\
  {\large \bf Thomas L. Griffiths$^{2,3}$ (tomg@princeton.edu)} \\
  {\large \bf Nathaniel D. Daw$^{1,2}$ (ndaw@princeton.edu)} \\
  $^1$Princeton Neuroscience Institute, Princeton University, Princeton, New Jersey \\
  $^2$Department of Psychology, Princeton University, Princeton, New Jersey \\
  $^3$Department of Computer Science, Princeton University, Princeton, New Jersey}

\begin{document}

\maketitle

\begin{abstract}
Typical models of learning assume incremental estimation of continuously-varying decision variables like expected rewards.
However, this class of models fails to capture more idiosyncratic, discrete heuristics and strategies that people and animals appear to exhibit.
Despite recent advances in strategy discovery using tools like recurrent networks that generalize the classic models, the resulting strategies are often onerous to interpret, making connections to cognition difficult to establish.
We use Bayesian program induction to discover strategies implemented by programs, letting the simplicity of strategies trade off against their effectiveness. 
Focusing on bandit tasks, we find strategies that are difficult or unexpected with classical incremental learning, like
asymmetric learning from rewarded and unrewarded trials,
adaptive horizon-dependent random exploration,
and discrete state switching.

\end{abstract}

Classic models of reinforcement learning (RL) often assume that humans and other animals incrementally estimate expected rewards or other continuously-varying decision variables, and make noisy decisions based on them \cite{rescorla1972theory,sutton2018reinforcement}.
However, there is increased recognition that the learning strategies actually adopted by humans (and even animals) deviate qualitatively from these models and instead reflect more discrete, idiosyncratic, and perhaps explicit heuristics: elaborating simple state-switching strategies like win-stay, lose-shift \cite{iigaya2018effect,lau2005dynamic} and reflecting memories of individual trials rather than summaries \cite{collins2012how,plonsky2015reliance,duncan2016memory,bornstein2017reminders}. 
Despite individual attempts to capture these behaviors by elaborating classic models, there is no general formalism spanning the space of such strategies or predicting what variants organisms adopt in particular situations, and how. 

Recent work on computational approaches to strategy discovery has started to engage with these questions.
For example, meta-RL has been used to train recurrent neural networks (RNNs) to have recurrent dynamics that implement a learning process \cite{duan2016rl2,wang2018prefrontal} which approximate decision-making strategies that are either task-optimized \cite{ortega2019metalearning} or supervised to distill organisms' learning behavior \cite{dezfouli2019models,ji_an2023automatic,miller2023cognitive}.
However, this approach is lacking in a number of respects.
First, neural networks require considerable effort to interpret (making them awkward for theory discovery) and are data- and compute-intensive to optimize (making them unlikely candidates for a process model of how organisms discover strategies). 
Relatedly, it is not clear that they represent the most appropriate strategy space over which to conduct meta-learning for these purposes. 
While the divergence of task-optimized RNNs from observed behavior might be addressed by introducing resource costs
\cite{moskovitz2023unified,binz2022modeling}, capacity-limited RNNs have low-dimensional continuous state spaces that do not capture the types of memory-dependent dynamics discussed above.

Accordingly, we propose instead to use Bayesian program induction to identify program-structured strategies for meta-learning in RL problems. This builds on approaches used to study cognitive representations \cite{goodman2008rational,yang2022one,ellis2022synthesizing} and discover machine learning algorithms \cite{real2020uto} that have been adapted to sequential decision making \cite{toussaint2006probabilistic,wingate2011bayesian}.
Our Bayesian framework has a natural interpretation as a resource-rational approach \cite{lieder2020resource} by letting the cognitive cost of a program correspond to its description length, so that simpler strategies are more probable under the prior but more effective strategies are more likely to result in optimal behavior under the likelihood.
Importantly, we can explore the relationship between previously-reported behavior and the strategies resulting from different trade-offs between effectiveness and program simplicity.
By focusing on program-structured strategies for RL, we explore a broader space of strategies than has previously been considered.
These programs are interpretable and, since they are small and can be extended with modularity and reuse \cite{yang2022one,ellis2021dreamcoder}, may also lead to a process-level model of strategy induction.

Using this approach, we identify a range of strategies in various bandit tasks, accounting for some previously observed behavioral phenomena that deviate from classical incremental learning.
We find instances where strategies focus exclusively on reward instead of omission, resembling behavioral signatures in rodents \cite{parker2016reward} and asymmetric learning more broadly \cite{palminteri2023choice}.
We show how this approach supports adaptive, horizon-specific randomness in exploration, as previously observed \cite{wilson2014humans}.
We also find strategies that switch between discrete decision states focused on either exploiting known options or exploring new options, as in previous studies \cite{ebitz2018exploration}.
Our framework provides an interpretable alternative to existing methods for strategy discovery, which we hope can provide new insights in other domains.

\section{Program-Based Strategy Induction}

\newcommand{\posterior}[1][]{p(\pi \mid \programOptimal#1)}
\newcommand{\likelihood}[1][\pi]{p(\programOptimal \mid #1)}
\newcommand{\prior}[1][\pi]{p(#1)}
\newcommand{\programOptimalVar}{\Omega}
\newcommand{\programOptimal}{\programOptimalVar = 1}

We formulate strategies as programs and use sampling-based Bayesian inference to find programs that are both simple and effective. We first introduce a formalism for tasks and strategies, then describe our framework for inference, and close with implementation details.

\subsection{Tasks and Strategies}

\newcommand\historyFull[1]{a_{1:#1},o_{1:#1},r_{1:#1}}
\newcommand\history[1]{h_{#1}}
\newcommand\transition[0]{p(o_{t+1},r_{t+1} \mid h_t, a_{t+1})}
\newcommand\agent[0]{p(a_{t+1} \mid \history{t})}
\newcommand\firstAction[0]{q_1}
\newcommand\return[1]{\E_{\history{#1}} \sum_{t=1}^{#1} r_t}

We study finite-horizon tasks where at every time step $t$ an agent takes an action $a_t$, makes an observation $o_t$, and receives a reward $r_t$, resulting over time in a history of agent-environment interactions $\history{t}=(\historyFull{t})$.
Observations may only provide partial information about the state of the task \cite{kaelbling1998planning},
so future observations and rewards depend on the full history of agent-environment interactions, $\transition$.
Agents are modeled as $\agent$, producing a task-specific distribution of histories $p(\history{t+1} \mid \history{t}) = \agent \transition$.

Strategies require performing two computations at every time step: updating internal representations to incorporate new observations and rewards, and using that updated internal representation to act. Formally, strategies generate an updated memory $m_{t+1}$, based on information from the previous trial (memory $m_t$, action $a_t$, observation $o_t$, and reward $r_t$), which we refer to as the state update function $f$, 
\begin{equation}\label{eq:state-update}
m_{t+1} = f(m_{t}, a_{t}, o_{t}, r_{t}).
\end{equation}
Choice behavior is driven by the updated memory $m_{t+1}$ and, for convenience, information from the previous trial, by way of a policy function $g$ that returns unnormalized log probabilities for each action,
\begin{align}\label{eq:action}
\log \agent &\propto g(m_{t+1}, a_{t}, o_{t}, r_{t}).
\end{align}
A completely specified strategy $\pi=(m_1, \firstAction, f, g)$
also requires an initial memory $m_1$ and unnormalized log policy $\log p(a_{1}) \propto \firstAction$ for the first time step. At subsequent time steps, memory is first updated with Eq.~\ref{eq:state-update} and then action is taken with Eq.~\ref{eq:action}.
Agents following a strategy are evaluated based on the expected sum of rewards they accrue over $T$ steps,
$$
V(\pi) = \return{T}.
$$

\subsection{Inference}

We use Bayesian program induction
to search the space of strategies
with a generative prior over strategies and a likelihood based on the effectiveness of the strategy.

\begin{table}[tb]
\begin{center} 
\caption{
Primitive operations for strategies.
}

\label{tab:dsl-primitives} 
\begin{tabular}{p{2.8cm}p{4.8cm}} 
\hline
Primitives    &  Description \\
\hline
Arithmetic, Logic \\
\hline
\texttt{0, \ldots, 49} & Integers from 0 to 49 (inclusive)\\
\texttt{+}, \texttt{*} & Addition, multiplication\\
\texttt{-}, \texttt{1/(x)} & Negation, multiplicative inverse\\
\texttt{<}, \texttt{==} & Less than, equals\\
\texttt{\&\&}, \texttt{||}, \texttt{!} & And, or, negation\\
\texttt{if(c,x,y)} & Returns \texttt{x} if condition \texttt{c} is true, \texttt{y} otherwise\\
\hline
Vectors \\
\hline
\texttt{vec\_full(x)} & A vector filled with the value $x$\\
\texttt{vec\_n(x1,\ldots,xn)} & A vector where the first $n$ entries are supplied and others are 0, e.g., \texttt{vec\_2(x,y)=[x,y,0,0]}\\
\texttt{v[i]} & Returns $\texttt{i}^{th}$ entry of \texttt{v}\\
\texttt{assign(v,i,x)} & Updated copy of \texttt{v}, with \texttt{v[i]=x} \\
\texttt{add\_assign(v,i,x)} & Updated copy of \texttt{v}, with \texttt{v[i]=v[i]+x} \\
\hline
Inputs \\
\hline
\texttt{prev\_action} & Previous action, $a_t$\\
\texttt{reward} & Previous reward, $r_t$ \\
\texttt{state} & Memory from previous trial $m_t$ for $f$ or current trial $m_{t+1}$ for $g$\\
\hline
Action probabilities \\
\hline
\texttt{logit(l)} & For two-action tasks, $l = \log \frac{p(a=0)}{p(a=1)}$\\
\texttt{softmax(w,v)} & Uses unnormalized log probabilities in $v$, scaled by $w$ \\
\texttt{action(a)} & Takes action $a$ \\
\texttt{argmax(v)} & Takes action with earliest, maximum value in $v$\\
\hline
\end{tabular} 
\end{center} 
\end{table}

Each component of a strategy is a program, a composition of the primitive operations defined in Table~\ref{tab:dsl-primitives}.
The primitives correspond to simple arithmetic, logic, and vectors, as well as primitives specific to decision-making, such as previous actions, rewards, and action distributions.
For simplicity, we let agent states be vectors of length 4.
The space of programs consists of all valid combinations of primitive operations that return the appropriate type (either a vector for $m_1$ and $f$, or action probabilities for $\firstAction$ and $g$), which corresponds to a context-free grammar.
The prior, $p(\pi)$, assumes expansions in the grammar are sampled uniformly at random, except terminals (like integers and inputs) are 8 times more likely than non-terminals in order to avoid excessive nesting. The integers, from 1 onward, are distributed according to a geometric distribution with probability $0.5$.
We represent actions with integers.
Invalid programs (non-integral or out of bounds actions/indices, computing $1/0$) were assigned a value of $-\infty$.
Programs that compute starting values ($m_1$, $\firstAction$) have a restricted set of primitives, excluding conditionals, vector indexing, and other input values.

Following the standard approach for planning by inference \cite{toussaint2006probabilistic,wingate2011bayesian,levine2018reinforcement}, we formulate our likelihood by introducing a Bernoulli-distributed random variable $\programOptimalVar$ that indicates the optimality of a strategy, so that
$$
\log \likelihood \propto \beta V(\pi),
$$
with $\beta$ a positive weight on the value.
Notably, optimal strategies maximize this likelihood.
Combining the likelihood and prior, the posterior probability that programs are optimal is
\begin{equation}\label{eq:posterior}
\log \posterior \propto \beta V(\pi) + \log \prior.
\end{equation}
The value weight $\beta$ controls the relative contribution of simplicity and task performance to the prior, so that simpler solutions are preferred when $\beta \to 0$ and more effective solutions are preferred when $\beta \to \infty$.

For inference, we use Markov chain Monte Carlo (MCMC) with Metropolis-Hastings acceptance to obtain samples from the posterior, as in prior work \cite{yang2022one,goodman2008rational}. We run five chains for $5\times10^5$ steps per weight $\beta$, which are $[100, 300, 1000, 3000, 10000, 30000]$ for the \citeA{wilson2014humans} task and $[10, 30, 100, 300, 1000, 3000]$ for others.
Proposal distributions are subtree-regeneration \cite{goodman2008rational}, and resampling a primitive (retaining arguments), swapping arguments, inserting a subtree between a parent and child, and deleting a subtree while retaining an ancestor \cite{yang2022one}.
A subset of ${m_1, q_1, f, g}$ was proposed to,
with each program independently sampled for inclusion (probability $\frac{1}{10}$ for $m_1$ and $q_1$ and $\frac{1}{5}$ for $f$ and $g$), avoiding empty subsets by always including a program sampled proportionally to inclusion probabilities.
Our implementation is built on the open-source Fleet library used in previous studies \cite{yang2022one}.

Value was estimated from $2 \times 10^3$ rollouts for the \citeA{ebitz2018exploration} task and $10^4$ rollouts in other cases, using expected immediate rewards as well as fixed environmental and agent seeds per chain to decrease variance.
To facilitate comparison across chains, we compare normalized value so that chance behavior corresponds to 0 and behavior guided by an oracle (i.e. complete knowledge of the reward probabilities) corresponds to 1.
Pareto frontiers were computed by finding the maximal strategies in the space defined by normalized value and the prior, using the the top 100 strategies from each chain.

\section{Exploring Strategies for a Simple Bandit Task}

\begin{figure}[tb!]
\centering

\begin{subfigure}[b]{.8\linewidth}
\centering
\caption{}
\label{fig:bandits:pareto} 
\includegraphics[width=\linewidth]{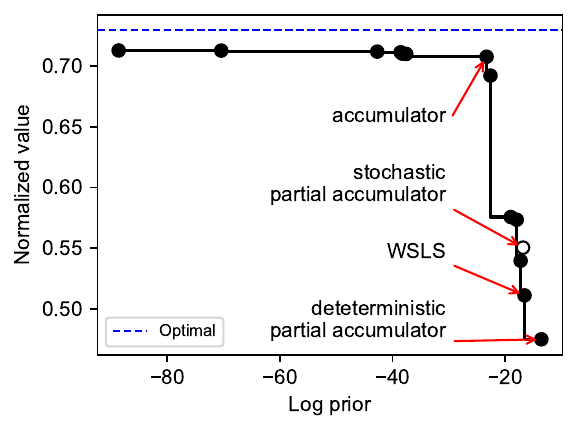}
\end{subfigure}%

\begin{subfigure}[t]{.5\linewidth}
\centering
\caption{}
\label{fig:bandits:wsls} 
\begin{tabular}{c}
WSLS \\
\begin{lstlisting}
m_1 = vec_2(0,1)
q_1 = action(0)
f = state
g = argmax(assign(
    state,prev_action,reward))
\end{lstlisting}\end{tabular}
\end{subfigure}%
\begin{subfigure}[t]{.5\linewidth}
\centering
\caption{}
\label{fig:bandits:accumulator} 
\begin{tabular}{c}
accumulator \\
\begin{lstlisting}
m_1 = vec_full(0)
q_1 = action(1)
f = add_assign(
    state,
    if(==(reward,prev_action),1,0),
    1
)
g = argmax(state)
\end{lstlisting}\end{tabular}
\end{subfigure}%

\begin{subfigure}[t]{.5\linewidth}
\centering
\caption{}
\label{fig:bandits:det-partial-accumulator} 
\begin{tabular}{c}
deterministic partial \\
accumulator \\
\begin{lstlisting}
m_1 = vec_full(0)
q_1 = action(1)
f = add_assign(
    state,prev_action,reward)
g = argmax(state)
\end{lstlisting}\end{tabular}
\end{subfigure}%
\begin{subfigure}[t]{.5\linewidth}
\centering
\caption{}
\label{fig:bandits:partial-accumulator} 
\begin{tabular}{c}
stochastic partial \\accumulator \\
\begin{lstlisting}
m_1 = vec_1(0)
q_1 = logit(0)
f = add_assign(
    state,prev_action,reward)
g = softmax(1,state)
\end{lstlisting}\end{tabular}
\end{subfigure}%

\caption{
Strategies for the two-armed bandit with stationary, Bernoulli rewards.
a)
The Pareto frontier of deterministic strategies, which are maximal points in the space defined by the normalized value and prior.
Points correspond to individual strategies.
One stochastic strategy described in the text is included (unfilled point) for comparison.
Value is normalized so that chance behavior has a value of 0 and behavior guided by an oracle has a value of 1.
Also shown is the Bayes-optimal solution to the task (dotted line).
Example strategies are shown in b-e), marked in a), and described in the text.
}
\label{fig:bandits} 
\end{figure}

We first survey the strategies discovered for a two-armed bandit, with a finite horizon of 20 trials and stationary, Bernoulli reward probabilities drawn from a uniform distribution.
The simplicity of this domain makes it possible to compute the value of strategies exactly. It is also possible to compare to the optimal solution, which has a simple analytic solution that can be reasonably computed for this horizon.

We first examine strategies when value is exactly computed, restricting attention to deterministic policies for computational reasons.
We focus on the Pareto frontier defined by simplicity and utility, that is, solutions that are local maxima in the space defined by simplicity and utility (Fig.~\ref{fig:bandits:pareto}).
Overall, the model exposes a progression of more elaborate and more effective strategies parameterized by this trade-off.

One simple solution we identify is win-stay, lose-shift (WSLS; Fig.~\ref{fig:bandits:wsls}),
where an action is repeated if the previous trial was rewarded, or an alternative action is selected if the previous trial was not rewarded \cite{robbins1952some}.
This solution has a trivial state update $f$ that simply passes a starting vector onward, with the strategy instead implemented entirely in the policy function $g$.
For the sake of clarity, we briefly walk through the evaluation of this strategy.
When \texttt{prev\_action=0} and \texttt{reward=1}, the result of \texttt{assign} is $[1, 1]$. Since \texttt{argmax} returns the earliest maximum value, the subsequent action is $0$, so winning results in repeating the action $0$.
On the other hand, if \texttt{reward=0}, the result of \texttt{assign} is $[0, 1]$, so \texttt{argmax} would result in the action $1$, meaning that a loss resulted in a switch.
The program behaves similarly for the other action.
While a more human-readable implementation might use a conditional, this implementation's careful use of indices avoids the need for a conditional and happens to be more probable under the prior.

A much more effective strategy accumulates the evidence indicated by rewards, so that wins increase the preference for an arm and losses decrease it. The algorithm in Fig.~\ref{fig:bandits:accumulator} implements this by incrementing the state at a conditional index, so that wins for an arm or losses for the other arm result in an increment as indexed by the arm. Since actions are selected by using \texttt{argmax}, the arm with the greatest accumulated evidence is selected at any time.

Other discovered strategies worse than the accumulator corresponded to partial accumulators that only incorporated part of the reward feedback in either an arm-specific or reward-specific way (like in Fig.~\ref{fig:bandits:det-partial-accumulator}). Strategies better than the accumulator were typically minor variants of it, like giving greater weight to outcomes for the arm that was not initially selected.

For validation, we examined the strategies discovered when value was estimated and stochastic policies were permitted.
We found similar policies as above, and use of stochastic policies that were often effectively deterministic due to large values of inverse temperatures.

One question is whether the resource-rational trade-offs implied by this model rationalize idiosyncratic behaviors in biological learning.
One notable random policy is the stochastic partial accumulator in Fig.~\ref{fig:bandits:partial-accumulator}, which shows a modest increase in value over WSLS.
Interestingly, this strategy ignores losses.
This can result in poor performance with deterministic behavior, since choice is fixed to the first rewarding option (Fig.~\ref{fig:bandits:det-partial-accumulator}, marked in Fig.~\ref{fig:bandits:pareto}). However, partial accumulation can perform effectively with stochastic choice.
This example is broadly consistent with findings of asymmetric learning from positive and negative prediction errors, such as exclusively learning from reward by rodents in a reversal learning task \cite{parker2016reward}.
It also echoes findings of optimistic learning biases in humans \cite{palminteri2023choice} that theoretical accounts (in a related task with counterfactual feedback) have shown to be normative in some settings \cite{lefebvre2022normative}.

To examine this more closely, we searched exhaustively over a small space of strategies to see whether exclusive accumulation of either rewarded or unrewarded trials was more effective.
The space of strategies was a generalization of the partial accumulators in Fig.~\ref{fig:bandits}, testing starting conditions, inverse temperature, and whether policies were deterministic or not.
When rewards were accumulated, the best strategy we found was Fig.~\ref{fig:bandits:partial-accumulator}. When unrewarded trials were accumulated (replacing \texttt{reward} with \texttt{+(-(1),reward)}), the best performing strategy was deterministic and implemented WSLS. Thus, within this limited class of strategies, a bias towards positive information is rational.

\begin{figure}[tb!]
\centering

\begin{subfigure}[t]{.5\linewidth}
\centering
\caption{}
\label{fig:wb:pareto} 
\includegraphics[width=.8\textwidth]{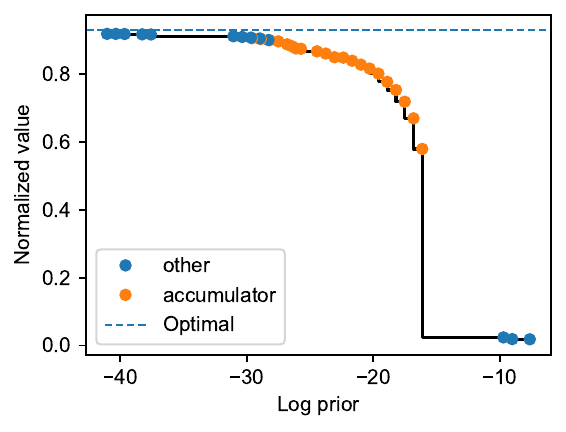}
\end{subfigure}%
\begin{subfigure}[t]{.5\linewidth}
\centering
\caption{}
\label{fig:wb:accum} 
\begin{tabular}{c}
stochastic accumulator \\
\begin{lstlisting}[escapeinside={(*}{*)}]
m_1 = vec_1(0)
q_1 = logit(0)
f = add_assign(
    state,prev_action,reward)
g = softmax((*\codePlaceholder*),state)
\end{lstlisting}\end{tabular}
\end{subfigure}%

\begin{subfigure}[t]{.5\linewidth}
\centering
\caption{}
\label{fig:wb:across-cond:original} 
\includegraphics[width=.95\textwidth]{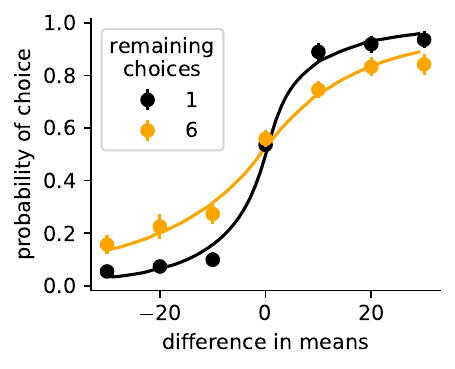}
\end{subfigure}%
\begin{subfigure}[t]{.5\linewidth}
\centering
\caption{}
\label{fig:wb:across-cond} 
\includegraphics[width=.95\textwidth]{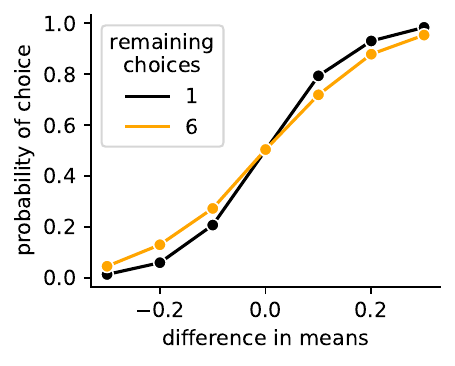}
\end{subfigure}%

\begin{subfigure}[t]{.5\linewidth}
\centering
\caption{}
\label{fig:wb:within-cond:original} 
\includegraphics[width=.95\textwidth]{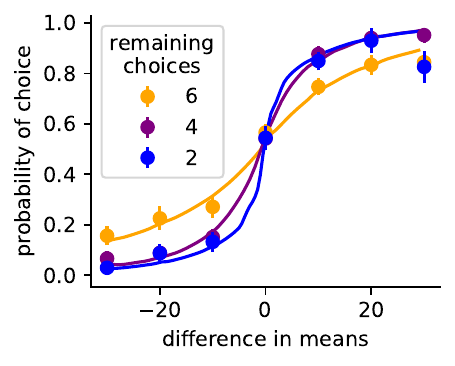}
\end{subfigure}%
\begin{subfigure}[t]{.5\linewidth}
\centering
\caption{}
\label{fig:wb:within-cond} 
\includegraphics[width=.95\textwidth]{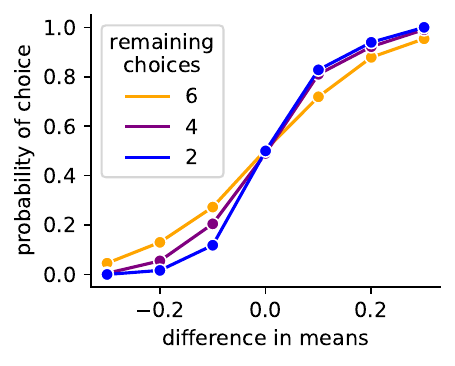}
\end{subfigure}%

\caption{
Adaptive random exploration, using a stochastic accumulator.
a) The Pareto frontier for solutions in the long horizon condition. Many solutions simply accumulate rewards, but increase the determinism of policies (less probable under prior) to achieve greater value.
b) The discovered accumulator. The ``\codePlaceholder'' is a placeholder for the inverse temperature.
c) Replotted from Fig.~2b in \protect\citeA{wilson2014humans}. Across conditions, there is a horizon-specific adjustment of decision noise.
d)
The optimal horizon-specific inverse temperature for the stochastic accumulator leads to more random behavior for the long horizon condition.
Horizon-specific inverse temperature was selected to maximize Eq.~\ref{eq:posterior} with $\beta=300$. Rewards were scaled by $\frac{1}{100}$.
e) Replotted from Fig.~3a in \protect\citeA{wilson2014humans}. In the long horizon condition, there is horizon-specific adjustment of decision noise.
f) Memory magnitude in the accumulator grows over time, resulting in less randomness for later trials in the long horizon condition.
Same inverse temperature as d).
}
\label{fig:wb} 
\end{figure}

\section{Adaptive Random Exploration}

A key feature of optimal behavior in bandit tasks is adaptive use of exploration\textemdash exploring is important early on, but over time behavior favors the best choice given previous observations.
\citeA{wilson2014humans} showed
that humans explore adaptively in a bandit task, with increased preference for unfamiliar and less valuable arms for a longer horizon and decreased preference for both for a shorter horizon.
While horizon-dependent exploration of unfamiliar arms is predicted by normative accounts, adaptive random exploration is hard to explain since optimal behavior is deterministic.

The task is a two-armed bandit with normal rewards that
starts with four forced-choice trials, to control the information participants have about each arm, then proceeds to either one or six free-choice trials.
Forced-choice trials ignored action probabilities, and an additional primitive was added so that strategies could test whether the previous trial was a forced-choice trial.
Strategies were found for each horizon condition independently and, to facilitate comparing conditions, value was the average received reward on free-choice trials.

Applying our framework to search for stochastic strategies for this task, we primarily identified an accumulator along the Pareto front (long horizon shown in Fig.~\ref{fig:wb:pareto}, short horizon was similar), with varied inverse temperatures for the \texttt{softmax} (Fig.~\ref{fig:wb:accum}, increasing with the value weight because the prior disfavors larger constants). This strategy performs best with a high inverse temperature, resulting in near-deterministic behavior.
In this section, we will use this example program to account for the adaptive random exploration found by \citeauthor{wilson2014humans}, focusing on the equal-information conditions where it is most directly exposed.

One pattern of adaptive random exploration was between different time horizons: on the first free-choice trial, participants were more random in the long horizon condition (Fig.~\ref{fig:wb:across-cond:original}).
Focusing on parametrically varying the strategy in Fig.~\ref{fig:wb:accum}, we found horizon-specific inverse temperatures for the \texttt{softmax} that maximized the posterior (Eq.~\ref{eq:posterior}). The inverse temperature was larger for the short horizon, resulting in more deterministic behavior compared to the long horizon (Fig.~\ref{fig:wb:across-cond}), consistent with the behavioral findings.

This horizon-dependent randomness results from an interplay between the accumulator's memory dynamics and the representation cost of the policy.
Accumulating rewards instead of weighting them means that the magnitude of entries in the agent's state will increase over time, leading to greater determinism. So, for the long horizon condition, the inverse temperature primarily serves to make early choices more deterministic.
Since integers are geometrically-distributed, larger values of the constant softmax temperature are less probable. This penalizes the determinism of a policy, resembling information-theoretic approaches to penalizing policy complexity \cite{piray2021linear,lai2021policy}.
Combined, this means that a strategy specific to the long horizon condition can afford to be more random early on, avoiding the penalty associated with being deterministic.

The second pattern of adaptive random exploration identified by \citeA{wilson2014humans} is within the long horizon condition: participant choice was more random when there were more remaining trials (Fig.~\ref{fig:wb:within-cond:original}).
While these results incorporated free-choice trials, they attempted to avoid a confound between information and reward by focusing on cases where each arm had been selected an equal number of times.
Using the best inverse temperature for the long horizon, we saw similar results (Fig.~\ref{fig:wb:within-cond}).
As described above, the accumulator's entries grow over time, resulting in a natural shift toward more deterministic behavior.

\begin{figure}[tb!]
\centering

\begin{subfigure}[t]{.5\linewidth}
\centering
\caption{}
\label{fig:ebitz:hmm} 
\includegraphics[width=\textwidth]{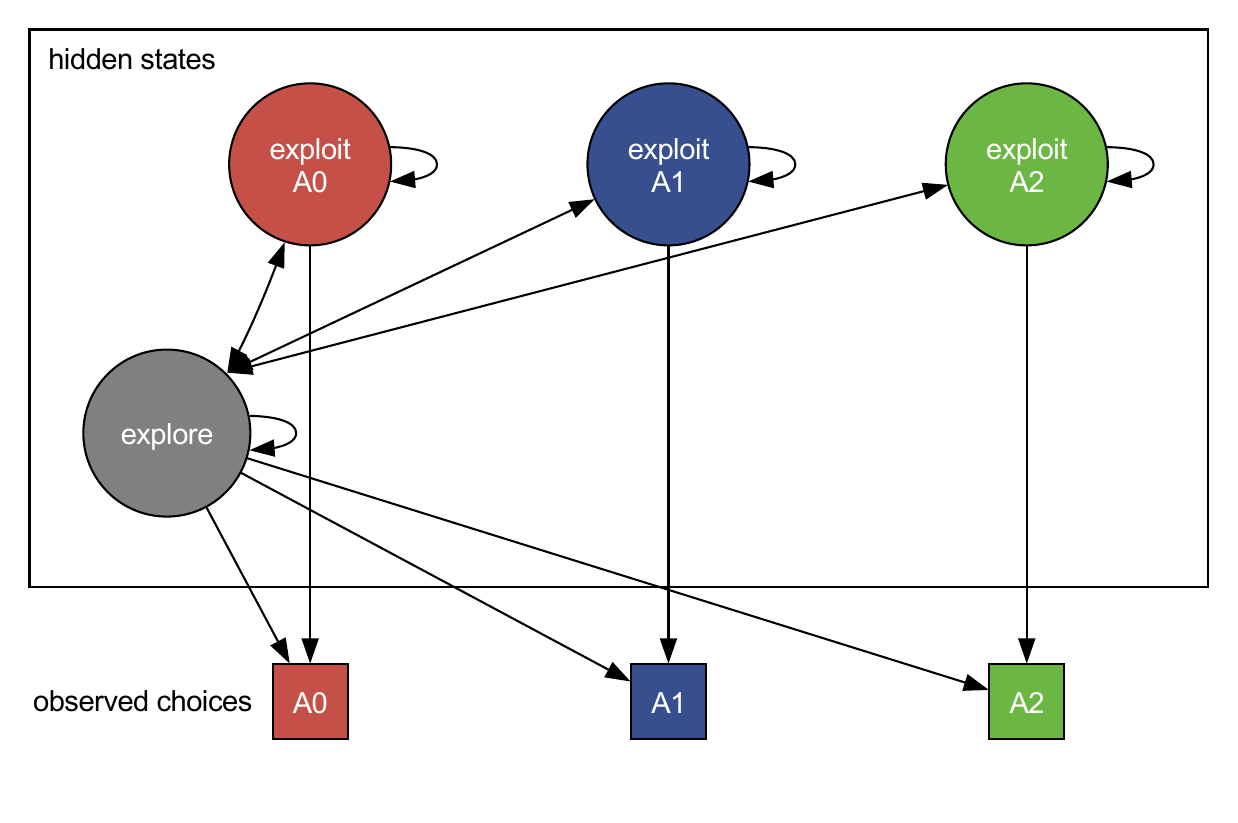}
\end{subfigure}%

\begin{subfigure}[c]{.55\linewidth}
\centering
\caption{}
\label{fig:ebitz:simple:code} 
\begin{tabular}{c}
\begin{lstlisting}
m_1 = vec_full(0)
q_1 = softmax(0,vec_full(0))
f = state
g = softmax(
    reward,
    assign(state,prev_action,7)
)
\end{lstlisting}\end{tabular}
\end{subfigure}%
\begin{subfigure}[c]{.4\linewidth}
\centering
\caption{}
\label{fig:ebitz:simple} 
\includegraphics[width=\textwidth]{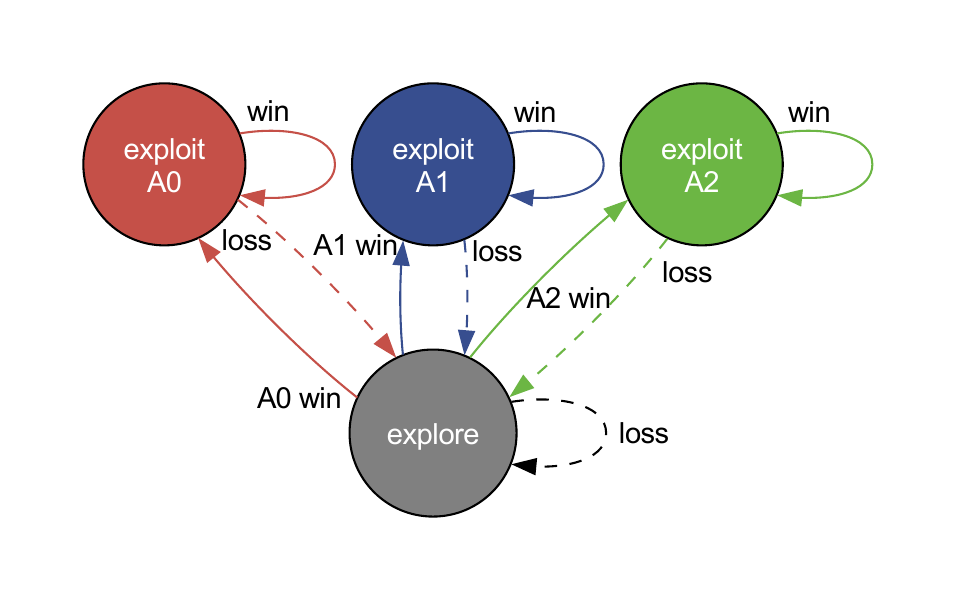}
\end{subfigure}%

\begin{subfigure}[c]{.55\linewidth}
\centering
\caption{}
\label{fig:ebitz:complex:code} 
\begin{tabular}{c}
\begin{lstlisting}
m_1 = vec_1(4)
q_1 = softmax(0,vec_1(0))
f = vec_1(
    +(1,state[reward]))
g = softmax(6, assign(
    vec_full(state[reward]),
    prev_action,
    4
))
\end{lstlisting}\end{tabular}
\end{subfigure}%
\begin{subfigure}[c]{.4\linewidth}
\centering
\caption{}
\label{fig:ebitz:complex} 
\includegraphics[width=.8\textwidth]{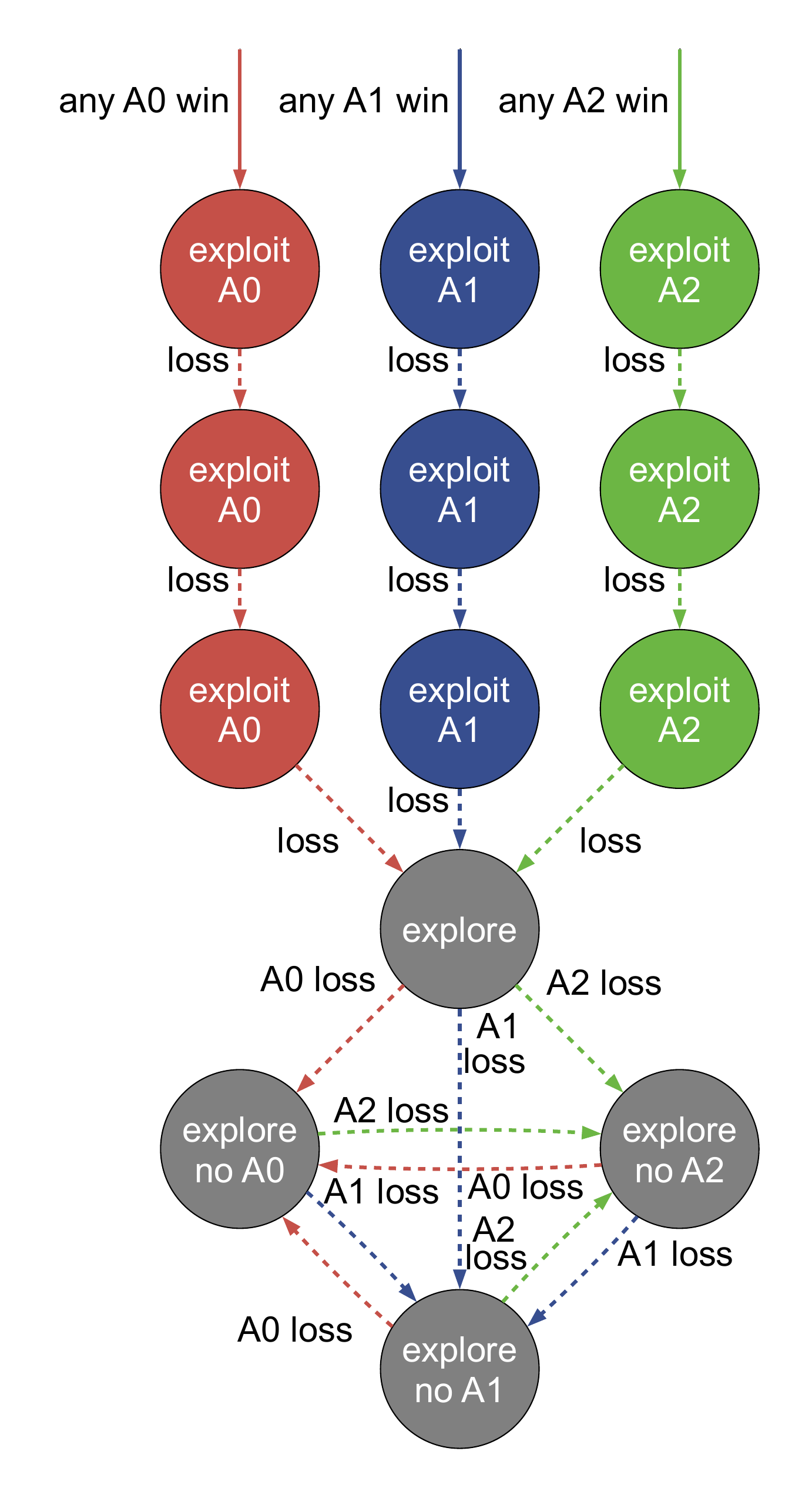}
\end{subfigure}%

\caption{
Discovering strategies with discrete decision states for a bandit task with non-stationary reward.
a) The state-based choice model in \protect\citeA{ebitz2018exploration}, featuring distinct states where behavior is either more exploratory or exploitative. Adapted from Fig.~1c in \protect\citeA{ebitz2018exploration}.
b) A WSLS strategy that randomly samples from actions after a loss.
c) A state machine for WSLS strategy.
States correspond to distinct action distributions. Edge color indicates action, dashed edges correspond to losses, and edges with action probability of less than 1\% were excluded.
The initial conditions of the strategy were modified to simplify the state machine.
d) A more complex strategy that exploits after a single win, but requires consecutive losses to switch.
e) State machine for d). Wins always lead to a state at top (as indicated), so wins are excluded elsewhere. States with 4+ consecutive losses (at bottom) were collapsed because they have similar action probabilities and identical transitions.
}
\label{fig:ebitz} 
\end{figure}

\section{Discrete Decision States}

We next examine strategies that emerge for learning in a bandit task with non-stationary rewards, which is difficult to solve with the accumulation models presented in previous sections.
\citeA{ebitz2018exploration} found that both choice behavior and neural activity from monkeys in this setting were captured by a slightly generalized WSLS model that involved switching between distinct states: exploring by random sampling, and exploiting by selecting a preferred action (Fig.~\ref{fig:ebitz:hmm}).
Quickly switching between several distinctive regimes of behavior is difficult to model using error-based learning.

We study the task reported by \citeauthor{ebitz2018exploration}, which is a 3-armed bandit with Bernoulli rewards. The probability of reward for each arm is in $(0.1, 0.2, \ldots, 0.9)$. After every trial, the probability has a 10\% chance of drifting up or down by $0.1$, while staying in bounds. We set the horizon to 500 trials.

Our framework identifies strategies with similar motifs as the one proposed by \citeauthor{ebitz2018exploration}, so we qualitatively examine two in this section.
The simpler program in Fig.~\ref{fig:ebitz:simple:code} implements WSLS where losses lead to random sampling among options\textemdash a loss means \texttt{reward=0}, so $\texttt{softmax}$ returns a uniform distribution. We also show the program's dynamics as a state machine in Fig.~\ref{fig:ebitz:simple}, where each state corresponds to a unique distribution over actions, and the arrows show how wins and losses change the policy.
In the state machine, the explore state results in random choice. Wins lead to an action-specific exploit state, and losses lead back to the explore state.
This strategy precisely mirrors the schematic proposed by \citeauthor{ebitz2018exploration} %
However, while their detailed results are consistent with state switching, they also show that both neural and behavioral measures somewhat violate the Markov conditional independence properties implied by their WSLS model, suggesting the actual state space is more complicated.
We explore one such strategy next.

The strategy in Fig.~\ref{fig:ebitz:complex:code} (state machine in Fig.~\ref{fig:ebitz:complex}) elaborates both the exploration and exploitation phases. In particular, when exploiting (top) it switches only after three successive losses; for exploration (bottom) it tries random arms until a win, but avoids the most recent choice.
The code implements this by counting consecutive losses in the $0^{th}$ entry of memory, resetting the counter to 1 following a win, and using the counter to inform choice probabilities.
Importantly, this strategy retains the discrete, state-like nature of the proposal in \citeauthor{ebitz2018exploration} %
Similar to the bandit algorithms discussed above, this strategy also responds more readily to positive feedback, which is also consistent with theoretical findings (in a related task) that choice-confirmation bias is adaptive when reward is non-stationary \cite{lefebvre2022normative}.

\section{Discussion}

We have proposed a framework for strategy induction, using inferential methods to identify strategies that are both simple and effective.
By searching for program-structured strategies, our approach is able to find strategy representations that are more interpretable than results from other approaches. We use this framework to examine strategies for RL tasks that trade off between simplicity and effectiveness.
We were able to discover strategies consistent with previously observed behavior, like asymmetric sensitivity to prediction errors, adaptive random exploration, and use of discrete decision states.

We were able to justify adaptive, horizon-dependent random exploration in our framework, as previously observed \cite{wilson2014humans}.
Within a fixed class of strategies, our framework can cast this adaptive behavior as the result of a resource-rational trade-off. Other approaches could explain this by penalizing the divergence of policies from a uniform distribution \cite{lai2021policy,piray2021linear,levine2018reinforcement}.
However, our approach goes beyond simple information-theoretic penalties on action distributions since it can identify structured, task-specific strategies and penalize them based on their description length.

One limitation of our current approach is that the primary cost of a program is its representational cost. This neglects any execution-related costs the program might incur, like memory use.
Penalizing trial-specific computation costs could encourage strategies that conditionally perform fewer computations in some trials, which, for example, could result in habits which can be adaptive under a speed-accuracy trade-off \cite{dezfouli2012habits}.
Future research could also adjust the primitive operations available to reflect cognitive resource limitations. For example, many kinds of memory are known to be subject to capacity limits or noise, like working memory \cite{collins2012how} and action values \cite{findling2019computational}.

Our approach is a computational-level account \cite{marr1982vision} of strategy induction, leaving open questions about the algorithms underlying strategy discovery.
Future work could examine the process-level bias induced by varying the number of samples taken in MCMC or for value estimation.
Another question is how library learning, which adds common subprograms as primitives \cite{ellis2021dreamcoder}, could accelerate inference.
Library learning offers a simple way to perform continual learning, which, by contrast, is an open  challenge for neural networks (e.g., requiring task-specific network dynamics to be identified and frozen to avoid catastrophic interference in research by \citeNP{duncker2020organizing}).

While we have focused on resource-rational strategies for learning in a handful of bandit tasks,
future research could apply this framework to identify interpretable strategies for sequential decision making in settings like planning, problem solving, and other learning tasks.
Future work could also apply this framework to discover strategies directly from behavioral data, as others have done with RNNs \cite{miller2023cognitive,ji_an2023automatic}.
We hope that the continued development of strategy discovery methods can accelerate our understanding of cognition by identifying novel features of behavior \cite{collins2020dichotomies} and avoid misinterpretation of behavioral signatures \cite{collins2012how,akam2015simple}.

\section{Acknowledgments}

We thank Evan M. Russek, Flora Bouchacourt, and Mark K. Ho for helpful discussions about early versions of this project.
This research was supported by grants from the
U.S. Army Research Office (ARO W911NF-16-1-0474, awarded to NDD),
National Institute of Mental Health (R01MH135587, awarded to NDD),
and the NOMIS foundation (awarded to TLG).
The funders had no role in study design, data collection and analysis, decision to publish, or preparation of the manuscript.

\bibliographystyle{apacite}

\setlength{\bibleftmargin}{.125in}
\setlength{\bibindent}{-\bibleftmargin}

\bibliography{references}

\end{document}